\documentclass[final]{cvpr}

\usepackage{times}
\usepackage{epsfig}
\usepackage{graphicx}
\usepackage{amsmath}
\usepackage{amssymb}
\usepackage{booktabs}
\usepackage{multirow}
\usepackage{makecell}
\usepackage{url}
\usepackage[table]{xcolor}
\usepackage{caption}
\usepackage{subcaption}
\usepackage{enumerate}
\usepackage{inconsolata}
\usepackage{diagbox}

\usepackage[export]{adjustbox}
\usepackage{float}
\usepackage{epsfig}
\usepackage{xparse} 
\usepackage{wrapfig} 
\usepackage{tabularx}
\usepackage{fancyhdr}
\usepackage[hang,flushmargin]{footmisc} 
\usepackage{pifont} 
\usepackage{boxedminipage}
\usepackage{framed}
\usepackage{soul}
\usepackage{xspace}
\usepackage{courier} 
\usepackage{colortbl}

\usepackage[pagebackref=true,breaklinks=true,colorlinks,bookmarks=false]{hyperref}

\def\cvprPaperID{****} 

\def\confYear{2023}

\begin{document}

\title{PILL:Plug Into LLM with Adapter Expert and Attention Gate}

\author{
	Fangyuan Zhang\textsuperscript{\rm 1}\quad\quad
        Tingting Liang\textsuperscript{\rm 1}\quad\quad
	Zhengyuan Wu\textsuperscript{\rm 1}\quad\quad
         Yuyu Yin\textsuperscript{\rm 1}\thanks{Yuyu Yin is the corresponding author}\quad\quad
	\\
	\normalsize\textsuperscript{\rm 1} School of Computer Science and Technology, Hangzhou Dianzi University, China.\\
}

\maketitle

\begin{abstract}
Due to the remarkable capabilities of powerful Large Language Models (LLMs) in effectively following instructions, there has been a growing number of assistants in the community to assist humans. Recently,  significant progress has been made in the development of Vision Language Models (VLMs), expanding the capabilities of LLMs and enabling them to execute more diverse instructions. However, it is foreseeable that models will likely need to handle tasks involving additional modalities such as speech, video, and others. This poses a particularly prominent challenge of dealing with the complexity of mixed modalities. To address this, we introduce a novel architecture called \textbf{PILL}: \textbf{P}lug \textbf{I}nto \textbf{LL}M with adapter expert and attention gate to better decouple these complex modalities and leverage efficient fine-tuning. We introduce two modules: Firstly, utilizing Mixture-of-Modality-Adapter-Expert to independently handle different modalities, enabling better adaptation to downstream tasks while preserving the expressive capability of the original model. Secondly, by introducing Modality-Attention-Gating, which enables adaptive control of the contribution of modality tokens to the overall representation. In addition, we have made improvements to the Adapter to enhance its learning and expressive capabilities. Experimental results demonstrate that our approach exhibits competitive performance compared to other mainstream methods for modality fusion. For researchers interested in our work, we provide free access to the code and models at \url{https://github.com/DsaltYfish/PILL}.
\end{abstract}


\section{Introduction}

The long-term goal of artificial intelligence is to enable it to utilize knowledge in a human-like manner for tasks such as reasoning, thinking, analysis, and decision-making. With the remarkable instruction-following ability and astonishing comprehension of human language exhibited by large language models~\cite{chatgpt, gpt3, flattent5}, Universal Visual Language Models (VLMs)~\cite{gpt4, chowdhery2022palm, kosmos1} have made significant progress in the field of AI's multimodal capabilities. In order to achieve modular processing of multimodal tasks, recent VLMs primarily employ visual prompt generators to provide visual information to large language models, showcasing impressive zero-shot performance across various visual tasks~\cite{Alayrac2022Flamingo, li2023blip2, instructblip}. 

However, despite having modality information aligned with language, the internal structure of the underlying large language models still lacks the ability to process entangled modalities, resulting in most VLMs still struggling to comprehend complex multimodal prompts. One straightforward approach is to fully fine-tune the large language model or utilize the Parameter-Efficient Fine-Tuning (PEFT) method to enable the LLMs to learn how to handle other modal information. Yet, this may result in the mixture of multiple modality information, leading to interference with the original knowledge learned by the LLM.

\begin{figure}
    \centering
    \includegraphics[width=1\linewidth]{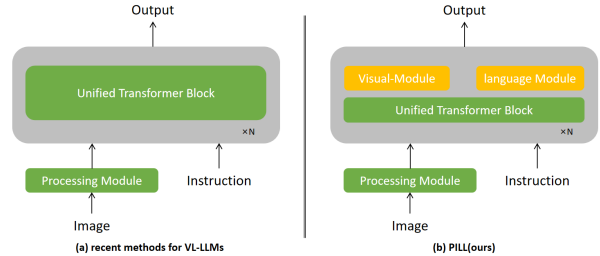}
    \caption{Comparison between recent VLMs methods and our approach is shown in Figure. (a) represents the recent VL-LLMs method. The Processing Module represents the modules used for image processing, including ViT, Qformer, Resampler, and others. These modules are employed to process image information and transfer it to the LLMs. (b) represents our approach. We keep the image processing modules consistent with other methods and introduce a dedicated module within the LLMs to handle image information.}
    \label{fig:compare}
\end{figure}

In this paper, we propose PILL: Plug Into LLM with Adapter Expert and Attention Gate, to address the aforementioned issues. Our method consists of three modules. Firstly, inspired by the Mixture-of-Modality-Experts (MoME) approach~\cite{vlmo, li2023pace}, we employ the Mixture-of-Modality-Adapter-Expert (MoMAE) to handle different modality inputs. In our experimental setup, MoMAE includes the V-Adapter and L-Adapter, which respectively handle the visual and textual modalities. Secondly, to ensure that the model produces results consistent with the original language model during initial training, we introduce the Modality-Attention-Gating (MAG) module. This stabilizes the training process and prevents visual information from interfering with the LLM's text modeling phase. Thirdly, recognizing that an enhanced adapter can lead to improved performance, we refine the classical adapter by introducing the GLU-Adapter.

To ensure stable training of our model, we adopt a two-stage training approach. In the first stage, we train the projection layer and V-Adapter separately to align visual and textual information, and enable the LLM to learn to process visual knowledge. In the second stage, we unlock all the proposed modules for downstream fine-tuning. Throughout both stages, the LLM and VPG module remain frozen. A notable advantage of our training method is its efficiency in training since we only train the inserted modules. Our method can be efficiently trained on a single A6000 GPU, allowing for rapid training completion.

The experimental results demonstrate that our model achieves competitive performance compared to other methods. On the ScienceQA dataset, our model achieves a 1.4\% accuracy improvement over models of equivalent scale. Compared to the best-performing LLaMA-based model, we achieve a 0.31\% accuracy improvement.

In summary, our contributions can be summarized as follows:

\begin{enumerate}
    \item We propose PILL, which addresses modality-specific information processing using MoMAE, dynamically adjusts modality injection into the LLM with MAG, and enhances performance with an improved adapter structure.
    \item Experimental results demonstrate that PILL exhibits superior efficiency and competitive performance compared to existing multimodal LLMs. It also showcases the significant potential for handling multimodal instructions while maintaining satisfactory performance on general VQA benchmarks.
    \item PILL demonstrates superior training and computational efficiency, as our experiments can be completed on a single A6000 GPU.
\end{enumerate}
\section{Related Work}
\subsection{Multi-Modal Pre-Training}
Large-scale pretraining has played a crucial role in the field of multimodal learning. Methods such as VL-BERT\cite{su2019vlbert}, ViL-BERT\cite{lu2019vilbert}, MCAN\cite{yu2019mcan}, LXMERT\cite{tan2019lxmert}, and Rosita\cite{cui2021rosita} extract key information from images using object detectors and process it together with text in transformer\cite{vaswani2017attention} blocks. Following the impressive success of CLIP\cite{radford2021clip} in image-text retrieval, contrastive learning has gained attention in the multimodal domain. Methods like ALBEF\cite{li2021albef}, BLIP\cite{blip}, VLMO\cite{vlmo}, BEiT-3\cite{wang2022beit3}, and CoCa\cite{yu2022coca} utilize contrastive learning to align image and text features, demonstrating the benefits of such alignment. Subsequently, with advancements in large-scale model techniques, methods like PaLi\cite{chen2022pali}, PaLM\cite{chowdhery2022palm}, KOSMOS\cite{kosmos1, peng2023kosmos2, lv2023kosmos2.5}, BLIP-2\cite{li2023blip2} build upon ClipCap\cite{mokady2021clipcap} by incorporating features from images specifically processed by ViT\cite{dosovitskiy2020vit} as prompts in the LM's input. More recently, with the rise of LLMs, researchers have focused on leveraging the powerful capabilities of LLMs and combining them with visual information. Mini-GPT4\cite{zhu2023minigpt} and LLaVA\cite{liu2023llava, liu2023llava1.5} have discovered that a projection layer can effectively project visual information into the textual space of LLMs, and they train only this projection layer using large-scale pretraining data.
\subsection{Multi-Modal Instruction-Tuning in PEFT}
To address the high cost associated with full fine-tuning, the Parameter-Efficient Fine-Tuning (PEFT) technique has emerged as an alternative. We highlight three commonly used PEFT techniques and discuss their applications in various methods. Firstly, the Adapter\cite{houlsby2019adapter} approach has been explored in VL-Adapter\cite{sung2022vladapter}. Extensive experiments have demonstrated the advantages of adapters over other PEFT methods in the multimodal setting. MAGMA\cite{eichenberg2021magma} introduces adapters within the LM based on a frozen model. LAVIN\cite{luo2023cheap} adopts a similar approach by treating single-modal and multimodal tasks as separate tasks and utilizing MMA (Mixture-of-Modality Adapter) for each task. The key distinction between our MoMAE approach and LAVIN is that we focus on modality tokens, while LAVIN focuses on tasks. Next, the application of LoRA\cite{hu2021lora} in the multimodal context is noteworthy. MultiModal-GPT\cite{gong2023multimodal-gpt} adopts a similar architecture to Flamingo\cite{Alayrac2022Flamingo} and incorporates LoRA for LM fine-tuning. Visual-ChatGLM\cite{du2022glm}, mPLUG-Owl\cite{ye2023mplugowl}, and LAMM\cite{yin2023lamm} also employ LoRA for LM fine-tuning after pretraining with VPG (Visual Prompt Generation). Finally, the prefix-tuning\cite{li2021prefixtuning} technique, exemplified by LLAMA-Adapter\cite{zhang2023llamaadapter, gao2023llamaadapterv2}, involves adding the image representation from the visual encoder to the prefix tokens and processing it along with the text tokens in the LM layers.
\begin{figure*}
\begin{center}
\includegraphics[width=1\linewidth]{figure/fig2.pdf}
\end{center}
   \caption{The overview of the architecture of PILL and two module of the PILL: Mixture-of-Modality-Adapter-Expert (MoMAE) and Modality-Attention-Gating (MAG). In PILL, the novel  MoMAE are employed to handle tokens from different modalities. During finetuning, MAG is used for coordinate the weights of other modalities}
\label{fig:framework}
\end{figure*}

\section{Method}
In this section, we will present the technical details of PILL, including the model architecture (Sec.\ref{sec:Architecture}), Mixture-of-Modality-Adapter-Expert (Sec.\ref{sec:MoMAE}) for handling different modalities of information, Modality-Attention-Gating (Sec.\ref{sec:MAG}) enables adaptive control of the contribution of modality tokens to the overall representation, and the SwiGLU-Adapter improves in adapter learning and expressive capability (Sec.\ref{sec:Adapter}). After a detailed introduction of these modules, we will proceed to discuss the training process and objectives of the model (Sec.\ref{sec:Training}). An overview of the PILL framework is depicted in Figure \ref{fig:framework}.

\subsection{Model Architecture}\label{sec:Architecture}
Given a sample containing a set of images and text, we first process each image in the image set using the Q-former module from BLIP-2 to extract image features.  Q-former takes a fixed number of $K$ learnable queries to interact with image features. We then project the dimension of these features to match the dimension of LLaMA using a projection layer, and we obtained $V_j = \{v_{j1}, ..., v_{jK}\}$, which represents the visual prompts for the $j$-th interleaved image. For the text, using the tokenizer within LLaMA, tokens are encoded into embeddings as $h_i$, which represents the $i$-th text token. Next, these encoded features are concatenated based on the positions of the original text and images in the sample and fed into PILL.The input features can be represented as $H = \{h_1, h_2, ..., v_{11}, ..., v_{1K}, ..., h_i, ..., v_{j1}, ..., v_{jK}, ..., h_N\}$. We define that $H^v = \{v_{11}, ..., v_{1K}, ..., v_{j1}, ..., v_{jK}\}$ and $H^t = \{h_1, h_2, ..., h_N\}$

\subsection{Mixture-of-Modality-Adapter-Expert}\label{sec:MoMAE}
We propose a universal multimodal module, named MoMAE, which is inspired by MoME~\cite{vlmo, li2023pace} and Adapter. MoMAE introduces Mixture-of-Modality-Adapter-Expert as an alternative solution to the standard Transformer feed-forward networks for encoding different modalities. As is shown in \ref{fig:framework}(b), after leveraging the output vectors from the previous layer and adapter-based multi-head self-attention (MSA), MoMAE are able to capture and process modality-specific information by switching to different modality adapter experts. We use vision adapter expert for encoding images $H^v$, and language adapter expert for encoding text $H^t$, which can be formulated as:

\begin{equation}
\begin{aligned}
    {H_{out}^v} &= Adapter_v({H_{in}^v})  \\
    {H_{out}^t} &= Adapter_t({H_{in}^t})
\end{aligned}
\end{equation}

Previous research has demonstrated that knowledge is preserved in the feed-forward networks (FFNs) within the Transformer. Therefore, we employ MoMAE to enable LLM to learn knowledge for handling visual features. Additionally, in our experiments, we observed that the variance of visual tokens is typically significantly larger than that of textual tokens. Such distribution inconsistency can potentially lead to bad training results. Employing a dedicated module to handle visual tokens may be a prudent choice. Furthermore, due to the adoption of the Adapter strategy for learning, the weights of the original LLaMA are retained, allowing the alignment of visual and textual modalities in the same representation space. The alignment has been proved to be crucial for multi-modal tasks in previous work.

\subsection{Modality-Attention-Gating}\label{sec:MAG}
Regarding modal fusion, both Prefix-Tuning and Cross-Attn methods have demonstrated excellent performance. In the experimental findings of Lynx~\cite{zeng2023matters}, Prefix-Tuning generally achieves better performance, while the Cross-Attn model may require more hyper-parameter searching to achieve superior results. Is it possible to combine the advantages of these two methods? To address this, we propose MAG: Modality-Attention-Gating. Specifically, we apply MAG to filter and adjust the attention weights of other modality information, aiming to allow the modal information to play its respective role in different transformer layers. 

We specify $G:\mathbb{R}^{d_{dim}} \rightarrow \mathbb{R}^{n_{heads}}$. Given vision modality tokens, our MAG module is computed as follows:

\begin{equation}
\begin{aligned}
    G_v(H) &= Tanh({G(H_{in}^v)})V(H_{in}^v)  \\
    G_t(H) &= V(H_{in}^t)  \\
    S(H) &= Softmax(\frac{Q(H_{in})K(H_{in})}{\sqrt{d_{head}}})  \\
    MAG(H) &= [G_v(H), G_t(H)] \odot S(H)
\end{aligned}
\end{equation}

In attention layer, $H$ is reshaped into $(n_{heads}, T, d_{head})$. $G_v(H)$.The role of $G_v(H)$ is to multiply the attention heads of each modality by a gating weight, which is initialized to 0. Note that for every visual token within the same layer, they undergo the same gate. This allows visual information to have distinct effects across different layers while filtering out irrelevant information.

Based on previous empirical experience in LM, modules closer to the input layer primarily focus on language modeling, while modules near the output layer are responsible for knowledge injection. The pure use of Prefix-Tuning might introduce interference in the language modeling stage due to the injection of other modal information, as observed in the experiments with Flamingo~\cite{Alayrac2022Flamingo}, where the gate weights near the input approach zero. On the other hand, the benefit of Prefix-Tuning lies in injecting different information at different transformer layers, effectively adapting the modal information at each layer. This may be a contributing factor to the generally higher performance of Prefix-Tuning compared to Cross-Attn. Therefore, based on the above assumptions, with the inclusion of the MAG and MoMAE module, the model architecture resembles Prefix-Tuning, while the details resemble Cross-Attn. In fact, our results of gating weight is similar to Flamingo~\cite{Alayrac2022Flamingo}

\subsection{SwiGLU-Adapter}\label{sec:Adapter}

\begin{figure}[t]
\begin{center}
\includegraphics[width=1\linewidth]{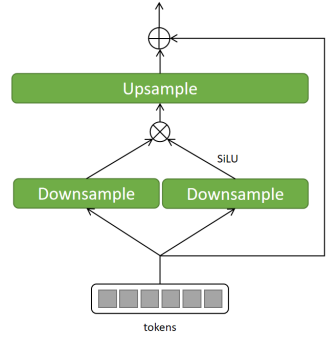}
\end{center}
   \caption{Architecture of SwiGLU-Adapter}
\label{fig:gluadapter}
\end{figure}

The classical Adapter architecture consists of a downsampling operation, activation function, upsampling, and residual connection. With the introduction of the Gated Linear Unit(GLU)~\cite{shazeer2020glu}  structure, it has gained wide application in LLMs such as LLaMA~\cite{touvron2023llama, touvron2023llama2} and PaLM~\cite{chowdhery2022palm}. In a corresponding manner, can we also transform the architecture of Adapter into GLU? To address this, we propose the SwiGLU-Adapter. As shown in \ref{fig:gluadapter}, we modify the structure of the Adapter by incorporating two downsampling operations, where one of the downsampling outputs is passed through an activation function. The outputs are then multiplied element-wise and followed by an upsampling operation. We represent this transformation using the following equation:

\begin{equation}
\begin{aligned}
    H_{down1} &= SiLU(D_1(H_{in}))  \\
    H_{down2} &= D_2(H_{in})  \\
    H_{down} &= H_{down1} \otimes H_{down2}  \\
    H_{up} &= U(H_{down})  \\
    H_{out} &= H_{up} \oplus  H_{in} \\
\end{aligned}
\end{equation}

\subsection{Stagewise Training}\label{sec:Training}
Our training process consists of the classic pre-training and downstream task fine-tuning stages, where different modules are trained at each stage. For training objective, both stage train on loss which is an auto-regressive manner. Given input $X_{text}$, $X_{image}$ and $X_{answer}$, we we express our training objective using the following equation:

\begin{equation}
\begin{aligned}
    &p_t=\prod_{s=1}^{S}p(X_s|X_{text}, X_{image}, X_{answer, <s};\theta)
\end{aligned}
\end{equation}

Here, $p_t$, $S$, $\theta$ denotes that the probabilities of the predicted word, the length of answer sequence, the trainable parameters, respectively. We maximize the likelihood of $p_t$ with trainable parameters $\theta$ in both training stage. The following sections will provide a detailed description of each stage.

\noindent \textbf{Stage 1: Pre-training for Modality Alignment.} 
In stage 1, similar to VLMO~\cite{vlmo, li2023pace} and LLaVA~\cite{liu2023llava}, we set $\theta = \{A_v, P\}$, where $A_v$ is visual adapter and $P$ is the projection matrix. By doing so, the visual features extracted from the Q-former can align with the pre-trained LLM word embeddings, allowing the FFN layers in LLM to learn preliminary knowledge in processing visual tokens. Note that the $A_v$ in the last layer does not have a connection to the training objective, rendering it untrainable. Therefore, we only utilize the V-Adapter as a T-Adapter in this case, effectively reversing the positional information of the input VL (Visual-Language) tokens.

\noindent \textbf{Stage 2: Downstream Task Fine-tuning.} 
In stage 2, we aim to train the model to respond to instructions and images, resembling a real AI assistant. Therefore, we train all the parameters of the Adapters, attn-gates, and projections, where $\theta=\{A_v, A_t, A_{attn}, G, P\}$. For quantitative experimental analysis, we conduct ablation experiments on the ScienceQA~\cite{lu2022learn}, a small-scale dataset. To achieve generalized instruction following, we train the model on the LLaVA1.5 665k~\cite{liu2023llava1.5} dataset.

\section{Experiment}

\begin{table*}[t]
\centering  
\resizebox{1.0\linewidth}{!}{
\begin{tabular}{lc|ccc|ccc|cc|c}  
\toprule
\multirow{2}{*}{Method} &\multirow{2}{*}{\#T-Param} & 
\multicolumn{3}{c|}{Subject} & \multicolumn{3}{c|}{Context Modality} & \multicolumn{2}{c|}{Grade} & 
\multirow{2}{*}{Average} \\
&& NAT   & SOC   & LAN   & TXT   & IMG   & NO    & G1-6  & G7-12\\ 
\midrule
\multicolumn{10}{l}{\it Zero- \& few-shot methods} \\   
Human~\cite{lu2022learn} &- &  90.23 & 84.97 & 87.48 & 89.60 & 87.50 & 88.10 & 91.59 & 82.42 & 88.40 \\
GPT-3.5~\cite{lu2022learn}&- &  74.64 & 69.74 & 76.00 & 74.44 & 67.28 & 77.42 & 76.80 & 68.89 & 73.97 \\
GPT-3.5 (CoT)~\cite{lu2022learn} &-  & 75.44 & 70.87 & 78.09 & 74.68 & 67.43 & 79.93 & 78.23 & 69.68 & 75.17 \\
GPT-4~\cite{gpt4} &- & 84.06 & 73.45 & 87.36 & 81.87 & 70.75 & 90.73 & 84.69 & 79.10 & 82.69 \\
\midrule
\multicolumn{10}{l}{\it Representative \&   SoTA  models}\\
UnifiedQA~\cite{lu2022learn}&223M& 71.00 &  76.04 &  78.91 &  66.42 &  66.53 & 81.81 &  77.06 & 68.82 &  74.11 \\
MM-CoT$_{Base}$~\cite{zhang2023multicot} &223M  &  87.52 & 77.17 & 85.82 & 87.88 & 82.90 & 86.83 & 84.65 & 85.37 & 84.91 \\
MM-CoT$_{Large}$~\cite{zhang2023multicot}& 738M  & 95.91 & 82.00 & 90.82 & 95.26 & 88.80 & 92.89 & 92.44 & 90.31 & 91.68 \\
\midrule
\multicolumn{10}{l}{\it LLaMA-based methods}\\ 
LLaMA-Adapter~\cite{zhang2023llamaadapter} &1.8M   & 84.37 & 88.30 & 84.36 & 83.72 &  80.32 &  86.90 &  85.83 &  84.05 & 85.19 \\
LaVIN~\cite{luo2023cheap} & 5.4M & 89.88 & 94.49 & \textbf{89.82} & 88.95 & 87.61 & \textbf{91.85} & \underline{91.45} & 89.72 & 90.83 \\
LLaVA~\cite{liu2023llava} &13B & \textbf{90.36} & \textbf{95.95} & 88.00 & \underline{89.49} & \underline{88.00} & 90.66 & 90.93 & \underline{90.90} & \underline{90.92} \\
LLaMA-SciTune~\cite{horawalavithana2023scitune} & 13B & 89.30 & 95.61 & 87.00 & \textbf{93.08} & 86.67 & \underline{91.75} & 84.37 & \textbf{91.30} & 90.03 \\
\rowcolor{lightgray}
PILL (ours) & 45M & \textbf{90.36} & \underline{95.84} & \underline{89.27} & 89.39 & \textbf{88.65} & 91.71 & \textbf{92.11} & 89.65 & \textbf{91.23} \\
\bottomrule
\end{tabular}
}
\vspace{2mm}
\caption{ Comparison on ScienceQA \textit{test} set.  Question classes: NAT = natural science, SOC = social science, LAN = language science, TXT = text context, IMG = image context, NO = no context, G1-6 = grades 1-6, G7-12 = grades 7-12. \#T-Params denotes that the number of trainable parameters.}  
\label{tab_sqa}  
\vspace{-1em}
\end{table*}

\subsection{Dataset}
\noindent \textbf{CC595K.} CC595K was filtered by LLaVA~\cite{liu2023llava, liu2023llava1.5}. LLaVa extracted noun phrases from each caption in the entire CC3M dataset and calculated the frequency of each unique noun phrase. Noun phrases with a frequency below 3 were skipped, as they typically represent rare combinations of concepts and attributes. For noun phrases with a frequency exceeding 100, a random subset of 100 samples was selected from all captions. This resulted in approximately 595k image-text pairs. We use CC595K to pretrain our model.

\noindent \textbf{ScienceQA.} ScienceQA~\cite{lu2022learn} consists of 21,000 data samples, including multiple-choice questions with multimodal content. It covers 3 subjects, 26 topics, 127 categories and 379 skills. We utilized the training split of the ScienceQA dataset, which comprised 12,726 samples, to further optimize our model. Additionally, we employed the test split, consisting of 4,241 samples, to conduct an initial evaluation of our model's performance.

\subsection{Implementation Details}
For the image feature extraction part, we utilized the Q-former module from BLIP2~\cite{li2023blip2}, which has been thoroughly pre-trained on Flan-T5-XXL~\cite{flattent5}. For the LLM, we opted for the LLaMA-2-chat~\cite{touvron2023llama2} model. The default setting for the intermediate hidden layer dimension in our adapter is 32. We employed AdamW as the optimizer with a cosine learning rate decay strategy.

During the pre-training phase, we solely trained the projection layer and v-adapter for 3 epochs, with a learning rate of 1e-3. The language model input length was set to 128, and the batch size was set to 32. In the fine-tuning stage, we trained the projection layer, v-adapter, t-adapter, attn-adapter, and attn-gate layer for 20 epochs, with a learning rate of 2e-3. The language model input length was increased to 512, and the batch size was reduced to 4.

Under our experimental settings, the total number of trainable parameters amounts to 45M. The training of our model was conducted on a single A6000, which is a hardware configuration widely acceptable among researchers.

\subsection{Experimental Results}
\subsubsection{Results on ScienceQA}
As shown in Table \ref{tab_sqa}, we compare the performance of several existing methods on the ScienceQA~\cite{lu2022learn}. Firstly, we consider zero-shot or few-shot approaches. Human performance in question answering achieves an accuracy of 88.40\%. Remarkably, GPT-4 achieves an accuracy of 82.69\% without any specific training on ScienceQA. Among other LLMs, MM-CoT$_{Large}$~\cite{zhang2023multicot} achieves an accuracy of 91.68\%. We speculate that this performance difference may be due to variations in the base models. Notably, there is a significant discrepancy in the NAT and SOC metrics between MM-COT and llama-based models. Furthermore, it is expected that MM-COT performs exceptionally well, as it focuses on the chain-of-thought, a technique that has been proven to significantly enhance LLM capabilities. Therefore, the outstanding performance of MM-COT comes as no surprise. Among the LLaMA-based methods, our method consistently ranks first or second across all metrics, ultimately achieving the best overall performance in terms of average metrics. Our method achieves comparable performance to the state-of-the-art methods, even without employing any extravagant techniques. In terms of parameter count, our method significantly reduces training overhead compared to LLaVA as we only need to train the adapter components of the model. Despite our method having significantly more training parameters compared to LaVIN, our training speed is faster due to the absence of gradient backpropagation to ViT during training. In fact, under the same experimental settings, our method requires only 90\% of the training time compared to LaVIN. Overall, these results validate the efficiency and design of PILL.

\begin{table}[t]
\centering
\resizebox{0.8\linewidth}{!}{
\begin{tabular}{l|cc}
\toprule
Settings                                   & w/o pretrain & w pretrain \\ \midrule

LLaMA-Adapter~\cite{zhang2023llamaadapter} & 85.19 & -\\
LLaVA~\cite{liu2023llava}                  & 85.81 & 89.84\\
LaVIN~\cite{luo2023cheap}                  & 89.41 & -\\
LLaMA-SciTune~\cite{horawalavithana2023scitune}  & - & 86.11\\
PILL(ours)                                 & \textbf{90.24} & \textbf{91.23}\\
\bottomrule
\end{tabular}}
\vspace{2mm}
\caption{ Results of existing multimodal LLMs in 7B model scaling on ScienceQA test set.}  
\label{tab:ablation_scienceQA_model_scale}
\vspace{-2em}
\end{table}

In Table \ref{tab:ablation_scienceQA_model_scale}, we compare the performance of PILL with other multimodal methods using the LLaMA-7B model specification. Without pretraining, LLaMA-Adapter\cite{zhang2023llamaadapter}, which was one of the pioneering methods to employ PEFT techniques, enabled the LLM to possess visual recognition capabilities and achieved impressive results on ScienceQA\cite{lu2022learn} at that time. LAVIN\cite{luo2023cheap} further improved upon this performance by approximately 4\%. We speculate that the Adapter structure may be particularly suitable for VL-LLM, as evidenced by the experimental results of VL-Adapter\cite{sung2022vladapter}. On the other hand, since LLaVA\cite{liu2023llava} requires full fine-tuning, a well-initialized visual projection layer is likely to stabilize the subsequent fine-tuning process. As a result, LLaVA achieves a 4\% improvement with pretraining. However, SciTune\cite{horawalavithana2023scitune}, despite its focus on analyzing figures and tables in scientific papers, performs poorly on some common-sense VQA tasks. Importantly, our method consistently achieves the best results regardless of whether pretraining is employed, demonstrating the effectiveness of our proposed approach.

\begin{figure}
    \centering
    \includegraphics[width=1\linewidth]{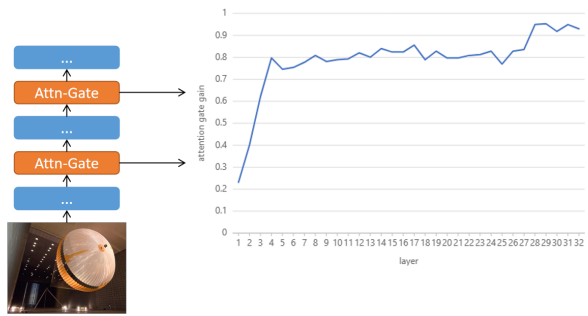}
    \caption{Evolution of the absolute value of the attention gate at different layers of PILL}
    \label{fig:attn-gate}
\end{figure}

As is shown in figure \ref{fig:attn-gate}, We also plotted the absolute values of the MAG at different layers of the PILL model, which consists of 32 LM layers. It appears that all layers of the frozen LM stack utilize visual information to some extent. We also observed that the absolute values tend to increase with depth. We attribute this to the roles played by the LLM at different layers. In the layers closer to the input, the LLM focuses on language modeling and does not heavily rely on visual information. However, as the layers get closer to the output, the LLM has accumulated sufficient knowledge and needs to extract information from the images. As a result, the absolute values of the tanh gate activations start from near-zero values near the input layers and rapidly increase, approaching 1 near the output layers.

\begin{table*}[t]\label{tab:ablation_scienceQA}
\centering
\resizebox{1.0\linewidth}{!}{
\begin{tabular}{lccccccccc}
\toprule
Settings                  & NAT   & SOC   & LAN   & TXT   & IMG   & NO    & G1-6  & G7-12 & \multicolumn{1}{l}{Avg.} \\ \midrule

\multicolumn{10}{l}{\it w/o pretrain}\\ 
MoMAE + L-A                                & 89.25 & 91.79 & 87.00 & 88.51 & 85.57 & 89.48 & 89.79 & 88.13 & 89.20 \tiny{\textcolor{red}{(+0.00)}}\\
MoMAE + MAG + L-A                          & 88.99 & 92.80 & 88.27 & 88.22  & 86.07 & 90.38 & 90.49 & 88.00 & 89.60 \tiny{\textcolor{red}{(+0.40)}}\\
MoMAE + MAG + G-A                          & 89.83 & 90.55 & 88.36 & 89.20 & 86.51 & 89.97 & 90.60 & 87.80 & 89.60 \tiny{\textcolor{red}{(+0.40)}}\\
MoMAE + MAG + SG-A                         & 90.01 & 93.36 & 88.18 & 89.15 & 87.16 & 90.59 & 91.12 & 88.66 & 90.24 \tiny{\textcolor{red}{(+1.04)}}\\
MoMAE + MAG + SG-A + 0.1 wrong answer      & 90.19 & 93.36 & 89.09 & 89.30 & 87.41 & 91.22 & 91.45 & 88.99 & 90.57 \tiny{\textcolor{red}{(+1.37)}}\\
\midrule \multicolumn{10}{l}{\it w pretrain epochs=1}\\ 
MAG + SG-A                                 & 90.36 & 94.38 & 87.45 & 89.83 & 88.10 & 89.83 & 90.90 & 89.65 & 90.45 \tiny{\textcolor{red}{(+1.25)}}\\
MoMAE + MAG + SG-A                         & 89.96 & 95.05 & 88.82 & 89.30 & 88.25 & 90.73 & 91.48 & 89.39 & 90.73 \tiny{\textcolor{red}{(+1.53)}}\\
MoMAE + MAG + SG-A + 0.1 wrong answer      & 89.43 & 95.61 & 89.09 & 88.66 & 87.85 & 91.01 & 91.34 & 89.39 & 90.64 \tiny{\textcolor{red}{(+1.44)}}\\
\midrule \multicolumn{10}{l}{\it w pretrain epochs=3}\\
MoMAE + MAG + SG-A                         & 90.36 & 95.84 & 89.36 & 89.93 & 88.80 & 91.15 & 92.33 & 89.32 & 91.23 \tiny{\textcolor{red}{(+2.05)}}\\
\bottomrule
\end{tabular}}
\vspace{2mm}
\caption{Ablation studies  on ScienceQA \textit{test} set. ``L-a'' denotes Linear Adapter. ``G-a'' denotes GELU Adapter. ``SG-A'' denotes SwiGLU Adapter. ``0.1 wrong answer'' indicates that during the training process, we randomly replace the ground truth answer with another option with a probability of 0.1.}  
\label{tab_ablation}
\vspace{-2em}
\end{table*}

\noindent \textbf{Ablation study.} In Table ~\ref{tab_ablation}, we present the contributions of various components and training processes of PILL on the ScienceQA test set. MoMAE and Linear Adapter serve as our baselines, achieving an accuracy of 89.20\%. Building upon this baseline, we incorporate the MAG module, resulting in a performance improvement of 0.4\%. To investigate the impact of different activation functions on the Adapter, we initially apply the GELU activation function, which does not lead to any improvement. However, when we replace the activation function with our proposed SwiGLU activation function, we achieve a further improvement of 0.64\%. Additionally, inspired by \cite{biten2022let}, we explore the effect of LLM hallucination on our model. During training, we randomly replace the answer with another option with a probability of 0.1, leading to an additional improvement of 0.33\%.

To further enhance PILL's ability to recognize images, we conducted image captioning training on the model. Initially, we attempted to remove the V-Adapter, training only a projection layer during pretraining and using a single adapter for fine-tuning. This approach achieved an accuracy of 90.45\%. When incorporating the MoMAE method, we observed an additional improvement of 0.28\%. We also explored the use of random wrong answers but encountered a decline in performance. We believe this is due to the lack of pretraining, which could result in insufficient image recognition capabilities and lead to hallucinations, similar to the findings in MM-COT\cite{zhang2023multicot}. After pretraining, the model exhibits improved image understanding, partially mitigating the hallucination issue. Therefore, employing the random wrong answer approach at this stage resulted in a decline in performance. Finally, by increasing the number of pretraining epochs to 3, we achieved the best results.

\section{Limitation}
Due to our utilization of LLM as the underlying model, PILL inevitably inherits some of the limitations of LLM. As a result, the image-to-text generation process may yield unsatisfactory results, including erroneous knowledge and reasoning or hallucinations. Additionally, although our work only involves the fusion of the image modality, our method can naturally incorporate other modalities such as speech. In future work, we aspire to create a genuine multimodal AI assistant by further exploring the integration of multiple modalities and scaling up the LLM.
\section{Conclusion}
In this work, we propose the PILL method to address the challenge of decoupling complex multimodal interactions. We leverage the MoMAE (Mixture-of-Modality-Adapter-Expert) module, which is specifically designed to handle image tokens, and the MAG (Modality-Attention-Gating) module for dynamic fusion of modalities. Additionally, we introduce the SwiGLU-Adapter to further enhance performance. Our experimental results demonstrate the effectiveness of our proposed method. With the advantage of fine-tuning only a small number of parameters, our approach offers a cost-effective solution that can be trained on a single A6000 GPU. This enables us to achieve visual language instruction following capabilities while maintaining efficient computation and training speed.

{\small
\bibliographystyle{ieee_fullname}
\bibliography{egbib}
}

\end{document}